# Does color modalities affect handwriting recognition? An empirical study on Persian handwritings using convolutional neural networks

Abbas Zohrevand[1], Zahra Imani[1], Javad Sadri[2,3*], Ching Y.Suen[4]

[1]Dept. of Computer Engineering, Kosar University of Bojnord, Bojnord, Iran.
{zohrevand86, z.imani13}@gmail.com

[2]Camp in Labs Foundation, 1 Westmount Square, Bureau 1001 Westmount (Québec) , Canada, H3Z 2P9

[3]Dept. of Computer Science and Software Engineering, Faculty of Engineering and Computer Science,  Concordia University, Montreal, Quebec, Canada.

j_sadri@encs.concordia.ca

[4]Center for Pattern Recognition and Machine Intelligence, Concordia University, Montreal H3G 1M8, Canada
suen@cs.concordia.ca

Abstract — Most of the methods on handwritten recognition in the literature are focused and evaluated on Black and White (BW) image databases. In this paper we try to answer a fundamental question in document recognition. Using Convolutional Neural Networks (CNNs), as eye simulator, we investigate to see whether color modalities of handwritten digits and words affect their recognition accuracy or speed? To the best of our knowledge, so far this question has not been answered due to the lack of handwritten databases that have all three color modalities of handwritings. To answer this question, we selected 13,330 isolated digits and 62,500 words from a novel Persian handwritten database, which have three different color modalities and are unique in term of size and variety. Our selected datasets are divided into training, validation, and testing sets. Afterwards, similar conventional CNN models are trained with the training samples. While the experimental results on the testing set show that CNN on the BW digit and word images has a higher performance compared to the other two color modalities, in general there are no significant differences for network accuracy in different color modalities. Also, comparisons of training times in three color modalities show that recognition of handwritten digits and words in BW images using CNN is much more efficient.

## 1. Introduction

Unlike the retina of some animals like cats and dogs which possess solely blue and green cones, humans have color vision due to the existence of red, blue and green cones in retina [1]. Color information plays an important role in human vision for recognition of elements such as shape, texture, and object recognition in daily life [2]. Similarly color as a rich feature and important descriptor is the vital part in many computer vision applications such as image super resolution [3], image segmentation [4], medical image analysis [5], remote sensing [6], image retrieval [7], and etc. Handwriting recognition is a very challenging task due to the existence of many difficulties such as the high variability of the handwritten styles and shapes, uncertainty of human-writings, skew or slant, segmentation of the words into isolated characters, and etc. [8]. Handwritten digit and word recognition has many applications such as processing of bank cheques [9], postal mail sorting[10], reading data from forms[11], notes recognition [12] and production of digital libraries of  historical documents [13]. While the effects of color as a feature has been studied in many computer vision applications, to the best of our knowledge so far no one has investigated the effects of color feature in handwritten recognition. This paper for first time tries to answer fundamental question in automatic document recognition to see whether color modalities of handwritten texts affect their recognition accuracy or speed?

---

[*] Corresponding Author

Handwriting recognition is the process of converting handwritten images into their corresponding editable files [8,14]. The handwritten recognition methods are divided into two main groups: on-line and off-line approaches. While on-line recognition depends on the trajectory of the pen and the coordinates of the pen's movement on the paper, off-line recognition is accomplished by analyzing the input image of the text [8]. Due to the important applications of handwriting in daily life, in the last twenty years many works have been proposed on handwritten digit and word recognition in different scripts such as Latin, Chinese, Indian, Arabic, and Persian. For example, see [15–18] for some recent works in isolated digit recognition. Similarly, Table 1 summarized some important works on the recognition of handwritten Persian/Arabic words. Although these works are impressive, most of them focus on BW images due to lack of handwritten digit and word images having all three color modalities. Fortunately, recently a new comprehensive Persian handwritten database which has all three color modalities (True Color (TC), Gray Level (GL), and Black and White (BW)) has been introduced in [19]. This paper consider samples of Persian isolated digits and word from this database and investigate to see whether color modalities of handwritten digits and words affect their recognition accuracy or speed?

Table 1. Some important works on the handwritten recognition in Persian/Arabic (BW: Black and White images)

| Ref. | Method | Dataset | Year | Color Modalities |
|------|--------|---------|------|------------------|
| [20] | Heuristic | IRANSHAHR | 2014 | BW |
| [21] | Image gradient, classifier fusion | IRANSHAHR | 2020 | BW |
| [22] | Fuzzy vector quantization, discrete HMM | IRANSHAHR | 2001 | BW |
| [23] | Structural based feature, discrete HMM, re-ranking | IFN/ENIT | 2011 | BW |
| [24] | M-band wavelet transform | Private | 2008 | BW |
| [25] | Statistical and structural features, discrete HMM | IFN/ENIT | 2018 | BW |
| [26] | Right to Left, Left to Right HMM | IRANSHAHR | 2018 | BW |
| [14] | Image gradient, discrete HMM | IRANSHAHR | 2014 | BW |
| [27] | Chain code, discrete HMM | IRANSHAHR | 2001 | BW |
| [28] | Statistical features, Support Vector Machine | IRANSHAHR | 2018 | BW |
| [29] | Connectionist Temporal Classification (CTC), CNN | Ref[19] | 2020 | BW |

Due to the complexity and variety of handwriting styles in different scripts, extracting the appropriate features from handwriting in any language is difficult and still open problem. In recent years, CNN models have been achieved considerable attention in the most computer vision applications (please see [4,30,31]). For the first time, CNN models proposed by LeCun et al [32,33]. These types of artificial neural networks combine feature extraction and classification tasks together, and the main purpose of their designs is to recognize image due to scale, shift and distortions [33]. Generally, CNNs consist of input and output layer as well as several convolutional layers sequentially connected to each other and followed by fully connected layer(s). In each convolutional layer inputs from the previous layer is convolved with trainable filters. Relu [34] usually acts as an activation function in CNN layers. For decreasing the size of the data and reducing the over-fitting phenomena [35], the pooling operation is applied to the output of the current layer. While CNNs as feature extractor or classifier have many applications in computer vision, so far they rarely have been used in Persian handwriting recognition. In this paper we try to use this architecture for recognizing Persian isolated digit and words in three color modalities.

To the best of our knowledge, so far, the effects of color modalities on handwriting recognition have not been studied. The main contribution of this paper is to answer this fundamental question. Using CNN, as an eye simulator, we investigated to see whether color modalities of handwritten isolated digits and words affect their recognition accuracy or speed. For our experiment, we adopted images of isolated digits and words from Persian database in [19] which have all three color modalities. In the first set of experiments, for isolated digit recognition, three similar CNN architectures are designed, trained, and tested. In the second set of experiments, a popular CNN architecture so called AlexNet [30] is fine-tuned for recognizing of handwritten words. In both set of experiments on digit and word recognition, results showed that Black and White (BW) color modality outperforms than other two color modalities (Gray Level (GL) and True Color (TC) level) in terms of recognition accuracy and training time of CNNs.

The rest of this paper organized as follow: in the Section 2 the database used for experiment are explained. In Section 3 the architecture of CNNs for all color modulates in isolated digits and words are explained. The details of experimental results on isolated digits and word in three color modalities are explained in Section 4, and finally in Section 5 conclusion and future works are described.

2.  Database

In the recent years besides developing many interesting and novel handwriting recognition methods, several handwritten databases in different scripts have been provided to evaluate and compare those recognition methods. For example for isolated digit, CENPARMI[†] digit dataset [36] which, comprises of 6000 digits prepared from the envelop USPS[‡] images. In this database 4000 samples, 400 images per digit, are determined for training and 2000 images for test. The CEDAR[§] digit dataset is another database, which comprises of 18468 and 2711 digits as the training and test sets, respectively [15]. The MNIST dataset [37] was another database which extracted from the NIST[**] dataset. In MNIST, images are normalized into 20 * 20 gray scale images, and then the normalized samples are placed in 28 * 28 frames [11]. Another popular database is USPS digit dataset which has 7291 training and 2007 test samples. Similar databases also exist for other languages such as Chinese, Arabic and Indian [38,39]. In Persian, HODA [11] is popular database which, their image samples were extracted from 12,000 registration forms. This dataset contains 102,352 digits and partitioned into train, test and evaluation groups. Similarly, Because of the existence of many challenges like the lack of certainty in human-writing, slant and skew in writing, styles and shapes with high variability, there are some valuable handwritten word databases in each script which summarized in Table 2. Although these datasets are impressive works, to the best of our knowledge most of them do not provide all color (TC, GL, and BW) modalities. Providing all color modalities of the same images of handwritten digits and words can provide a lot of new opportunities for investigation in handwriting recognition.

---

[†] Center for Pattern Recognition and Machine Intelligence.
[‡] United States Postal Service.
[§] Center of Excellence for Document Analysis and Recognition.
[**] National Institute of Standards and Technology.

Table 2. Some handwriting word recognition databases (TC: True Color, GL: Gray Level, BW: Black and White image modalities)

| Database Name | Script | No. of Writers | Words | Color Modalities |
|---|---|---|---|---|
| **CEDAR**[40] | English | 311 | 10,570 | BW |
| **GRUHD** [41] | Greek | 1000 | 102,692 | BW |
| **IAM**[42] | English | 780 | 82,227 | BW |
| **CENPARMI** [43] | Arabic | 328 | 11,375 | BW |
| **CENPARMI**[44] | Dari | 200 | 14,600 | BW |
| **CENPARMI**[45] | Persian | 175 | 8575 | BW |
| **IAUT/PHCN** [46] | Persian | 380 | 34200 | BW |
| **IFN/Farsi-database**[47] | Persian | 600 | 7271 | BW |
| **CENPARMI**[48] | Persian | 400 | 516 | BW |
| **IFN/ENIT**[49] | Arabic | 400 | 26,000 | BW |
| **IRANSHAHR**[50] | Persian | 503 | 15.090 | BW |
| **FARSA**[51] | Persian | 300 | 30,000 | BW, GL |
| **Sadri and et al**[19] | Persian | 500 | 62,500 | BW,GL,TC |

Recently, a newly Persian database [19] has been developed which consist of a huge number of numeral strings, touching digits, digits, numeral dates and worded, common financial words, Persian alphabetical letters, symbols, punctuation marks and an unconstrained text. This database was collected from 500 Persian male and female writers equally. Each of the 500 participants wrote 125 different words. For each stored image detailed ground truth files also were created. All sample images were recorded in three modalities: TC, GL, and BW. One example of three image color modalities in digits and words is shown in Table 3 and Table 4 respectively. In this paper, we use the samples of Persian isolated digits and words from this database in three color modalities for our experiments.

Table 3. Persian handwritten isolated digits samples [19][52] (TC: True Color, GL: Gray Level, BW: Black and White image modalities).

| Digit | 0 | 1 | 2 | 3 | 4 | 5 | 6 | 7 | 8 | 9 |
|---|---|---|---|---|---|---|---|---|---|---|
| TC | ۰ | ۱ | ۲ | ۳ | ۴ | ۵ | ۶ | ۷ | ۸ | ۹ |
| GL | ۰ | ۱ | ۲ | ۳ | ۴ | ۵ | ۶ | ۷ | ۸ | ۹ |
| BW | ۰ | ۱ | ۲ | ۳ | ۴ | ۵ | ۶ | ۷ | ۸ | ۹ |

Table 4. The samples of Persian handwritten word [19] (TC: True Color, GL: Gray Level, BW: Black and White image modalities).

| Words | Writer #1 | | | Writer #2 | | |
|---|---|---|---|---|---|---|
| | TC | GL | BW | TC | GL | BW |
| **Gentleman** | محترم | محترم | محترم | محترم | محترم | محترم |
| **Ramadan** | رمضان | رمضان | رمضان | رمضان | رمضان | رمضان |

## 3. Proposed methodology

Using CNN models, the present study investigate the effects of color modalities on handwritten digit and word recognition. In the following subsections CNN architectures for isolated digit and word recognition are explained.

### 3.1. Architecture for isolated handwritten digit recognition

The proposed CNN architecture for isolated digit recognition shown in Figure 1. This CNN network contains: one input layer, two convolutional, two sub-sampling layers for extracting features automatically, two fully connected layers as multi-layer perceptron for classification, and one output layer. Details of the proposed architecture shown in Table 5. Existence many trainable parameters are one of the challenging point in CNN [53]. As shown in Table 5, there are many parameters (in total 2,117,962) that should be optimized in CNN, and for training these huge numbers of parameters there must be enough image samples. In conventional CNNs, the first layers capture primary features of an input such as lines and edges. While the latter layers extract more complex parts of an input like tail of dog in an image [29]. As a result, the early layers converged after few epochs. Nevertheless, deeper layers need considerable epochs to converge. So, the neural networks are robust against most of the geometrical transformation [54]. As a consequence, when the network needs more data for training, it is easily possible augment new data by applying some simple geometrical transformations on the input data [30].

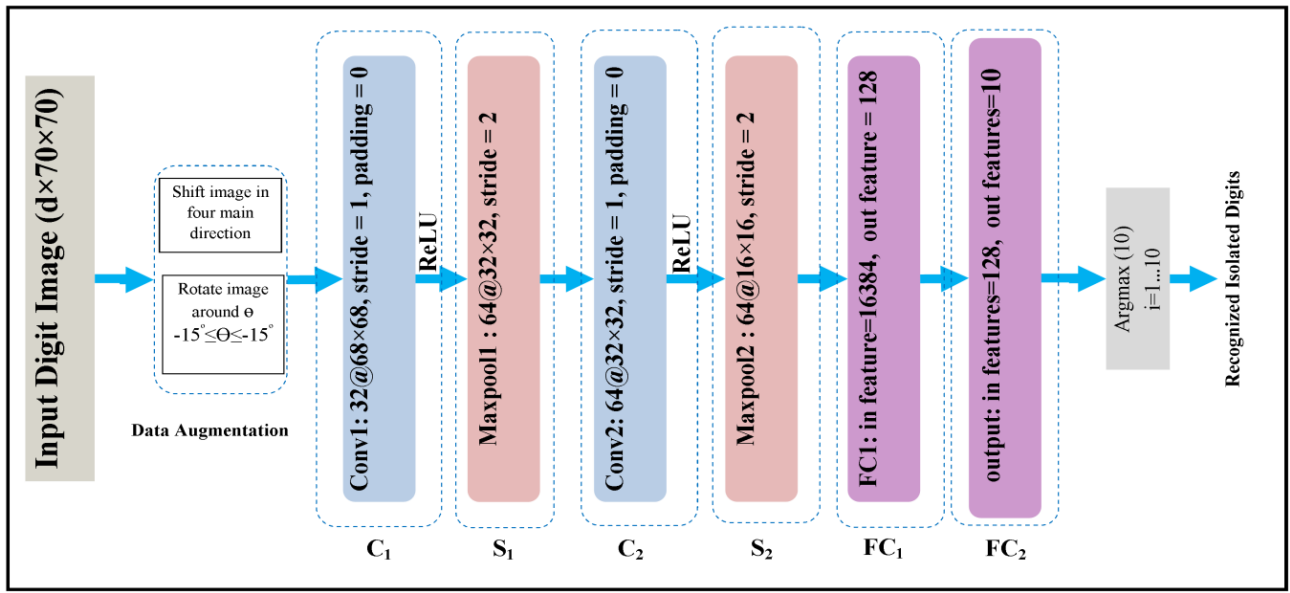

Fig. 1. Convolutional Neural Network[52] ($C_i$: convolutional layer, $S_i$: max-pool layer and $FC_i$: fully connected layer). In the input digit image 'd' indicates depth of the input images (d=1 for BW and GL images, d=3 for TC images)

Table 5. Detailes of each layer of proposed CNN[52] (shown in Figure 1).

| Layer | Layer Operation | Feature Map No. | Feature Map Size | Window Size | Parameter |
|---|---|---|---|---|---|
| C1 | Convolution | 32 | 68x68 | 3x3 | 896 |
| S1 | Max-pooling | 32 | 34x34 | 2x2 | 0 |
| C2 | Convolution | 64 | 32x32 | 3x3 | 18496 |
| S2 | Max-pooling | 64 | 16x16 | 2x2 | 0 |
| FC | Flatten layer | N/A | N/A | N/A | 0 |
| FC | Fully connected | N/A | N/A | N/A | 2097280 |
| FC | Output | N/A | N/A | N/A | 1290 |
| | | | | **Total =** | **2,117,962** |

In this paper, a subset including of 13,330 samples of isolated digits have been selected from database [19], then 80% of them have been selected for training and validating CNN models, and 20% remaining images for testing of trained CNN models. For creating enough images, we have used efficient simple geometrical transformation like shifting (in four main directions) and rotation (with constant probability). As shown in Figure 2, each input image firstly shifted in four main directions and then each shifted and original images are rotated around -15 to 15 degrees with constant probability. Totally after augmentation, there are 100,000 images, which 80% (0.8*100,000 = 80,000) of data provided for training and validation of CNN model and remaining 20%(0.2*100,000 = 20,000) are used for testing CNN model. We use the same architecture for recognition of isolated digit in TC, GL and BW color modalities, and then we compare these networks in terms of their recognition accuracies and their training times.

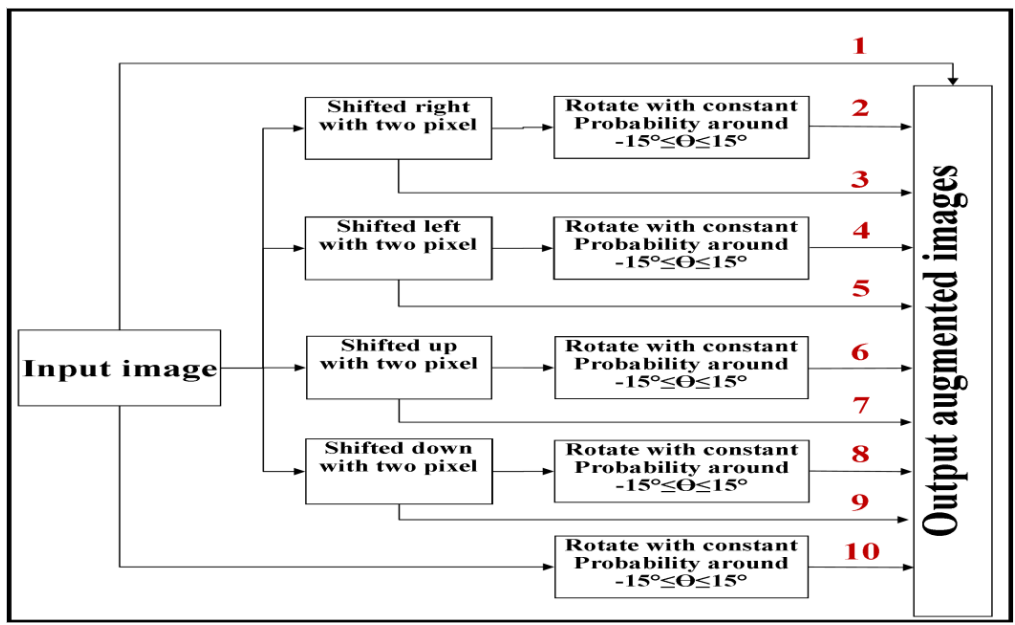

Fig. 2. Structure of augmenting for each input image[52]. As shown in the figure, input images and all their shifted are saved as new images, and then each shifted and original images are rotated randomly with a fix probability (= 0.9).

## 3.2. Architecture for handwritten word recognition

Automatic Persian handwritten word recognition is a very complex task due to the existence of many challenges like lack of certainty in human writing, styles and shapes with high variability in handwritten and etc. Segmentation-based [55] and holistic [56] are two main approaches in Persian word recognition. This paper focuses on holistic approach. Hinton et al in 2012 proposed new architecture similar to LeNet [37] but more complex and deeper so called "AlexNet" [30] (please see Fig 3). As shown in this figure, it contains five convolutional layers and three fully connected layers. Relu [34] is applied after every convolutional and fully connected layer. In this paper, we use this architecture as a backbone network, and then we fine-tune it by modified the three last layers for handwritten word recognition.

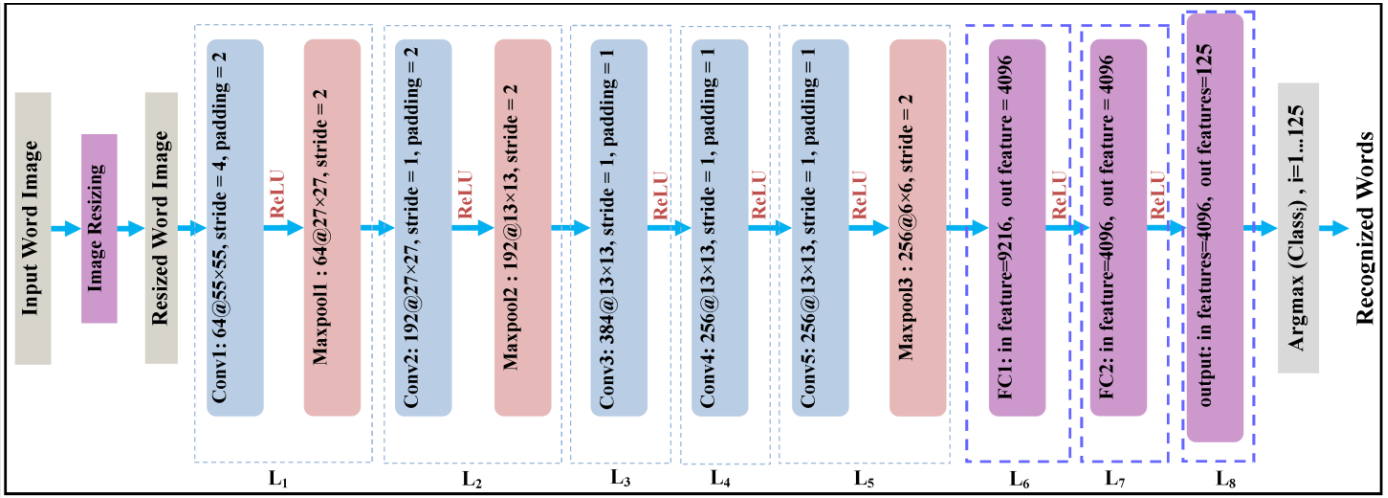

Fig 3. AlexNet as a backbone network. In this architecture, there are eight layers including five convolution layers($L_1 \ldots L_5$) and three fully connected ($L_6$, $L_7$, and $L_8$) layers. In total, there are 57,515,965 parameters in this network.

Training CNN from scratch without reducing the generalizability is not easy and requires plentiful training data. In conventional CNNs, the first layers capture global features of input image, while the last layers extract more local featurs from input images [30,54]. In this paper for fine-tuning a backbone network, only the three last fully connected layers ($L_6$, $L_7$, and $L_8$) are trained on handwritten word images. The general setup for fine-tuning procedure has shown in Table 6. As shown in this table, each fully connected layer has input features (in) and output features (out). The backbone network is trained with different numbers of hidden neurons in each of its three last layers ($L_6$, $L_7$ and $L_8$) and the network with best performance on BW word images are chosen. The detail of the best fine-tuned network is shown in Table 7. As shown in this table, the total number of parameter in the best fine-tuned network is equal to 12,195,529 which compared to AlexNet (57,515, 965) shown in Fig 3 is much lower, while keeping same performance. We use the best fine-tuned network to investigate the effects of color modalities on handwritten word recognition in the experimental results.

Table 6: The setup for fine–tuning AlexNet, in each row one or more than one fully connected layer are trained with special batch size, epochs and learning rate on BW images. Then the network with respect to the best setup is chosen.

| No. of trainable Layer | | | Batch Size | No. of Epochs | Learning Rate ($\eta$) | Accuracy (%) (BW) |
|---|---|---|---|---|---|---|
| $L_6$(in, out) | $L_7$(in, out) | $L_8$(in, out) | | | | |
| (9216,4096) | (4096,9096) | (1024,125) | 128 | 100 | 0.001 | 77.0 |
| (9216,4096) | (4096,4096) | (4096,125) | 128 | 100 | 0.01 | 69.0 |
| (9216,4096) | (4096,4096) | (4096,125) | 200 | 100 | 0.001 | 76.0 |
| (9216,4096) | (4096,250) | (250,125) | 128 | 100 | 0.001 | 89.6 |
| (9216,4096) | (4096,500) | (500,125) | 128 | 100 | 0.001 | 89.8 |
| (9216,4096) | (4096,700) | (700,125) | 128 | 100 | 0.001 | 85.0 |
| (9216,1024) | (1024,250) | (250,125) | 128 | 100 | 0.001 | 94.9 |
| (9216,1024) | (1024,500) | (500,125) | 128 | 100 | 0.001 | 95.0 |
| (9216,1024) | (1024,250) | (250,125) | 128 | 100 | 0.002 | 95.2 |
| (9216,**1024**) | (**1024**,**250**) | (**250**,125) | 128 | 100 | 0.004 | 95.4 |

Table 7: Details of the best fine-tuned network from Table 6.

| Layer name | Layer type | No. of filter | Kernel size | Stride | Padding | Input features | Output features | No. of parameters |
|---|---|---|---|---|---|---|---|---|
| | Convolution | 64 | (11,11) | (4,4) | (2,2) | (d,224,224) | (64,55,55) | 23,296 |
| | Relu | --- | --- | --- | --- | (64,55,55) | (64,55,55) | 0 |
| $L_1$ | Max-pooling | --- | (5,5) | (2,2) | --- | (64,55,55) | (64,27,27) | 0 |
| | Convolution | 192 | (5,5) | (2,2) | (2,2) | (64,27,27) | (192,27,27) | 307,392 |
| | Relu | --- | --- | --- | --- | (192,27,27) | (192,27,27) | 0 |
| $L_2$ | Max-pooling | --- | (3,3) | (2,2) | --- | (192,27,27) | (192,13,13) | 0 |
| | Convolution | 384 | (3,3) | (1,1) | (1,1) | (192,13,13) | (384,13,13) | 663,936 |
| $L_3$ | Relu | --- | --- | --- | --- | (384,13,13) | (384,13,13) | 0 |
| | Convolution | 256 | (3,3) | (1,1) | (1,1) | (384,13,13) | (256,13,13) | 884,992 |
| $L_4$ | Relu | --- | --- | --- | --- | (256,13,13) | (256,13,13) | 0 |
| | Convolution | 256 | (3,3) | (1,1) | (1,1) | (256,13,13) | (256,13,13) | 590,080 |
| $L_5$ | Relu | --- | --- | --- | --- | (256,13,13) | (256,13,13) | 0 |
| | Max-pooling | --- | (3,3) | (2,2) | --- | (256,6,6) | (256,6,6) | 0 |
| | Fully connected | --- | --- | --- | --- | 9216 | 1024 | 9,438,208 |
| $L_6$ | Relu | --- | --- | --- | --- | --- | --- | 0 |
| | Fully connected | --- | --- | --- | --- | 1024 | 250 | 256,250 |
| $L_7$ | Relu | --- | --- | --- | --- | --- | --- | 0 |
| | Fully connected | --- | --- | --- | --- | 250 | 125 | 31,375 |
| $L_8$ | Relu | --- | --- | --- | --- | --- | --- | 0 |
| | | | | | | | **Total** = | **12,195,529** |

Database in [19] have word images with different sizes, however AlexNet [30] need 227*227 image as input size in first layer. There are many approach for resizing images [19]. The simplest method is resizing all images to 227*227, but applying this method on all images with different sizes deal to deform structure of handwritten images. In this paper, motivated by [57], the efficient image scaling (resizing) applied on all images in the preprocessing step (please see Figure 4). As shown in this figure, all images in database divided into two groups; firstly, images with dimensions smaller than standard size. These images only are padded with background pixel. Secondly, images with one or both dimension greater than 227 sizes. In these images, the ratio of the standard size to the bigger dimension is calculated and height and width of the input image is resized base on this ratio. With this strategy structure of input images doesn't change.

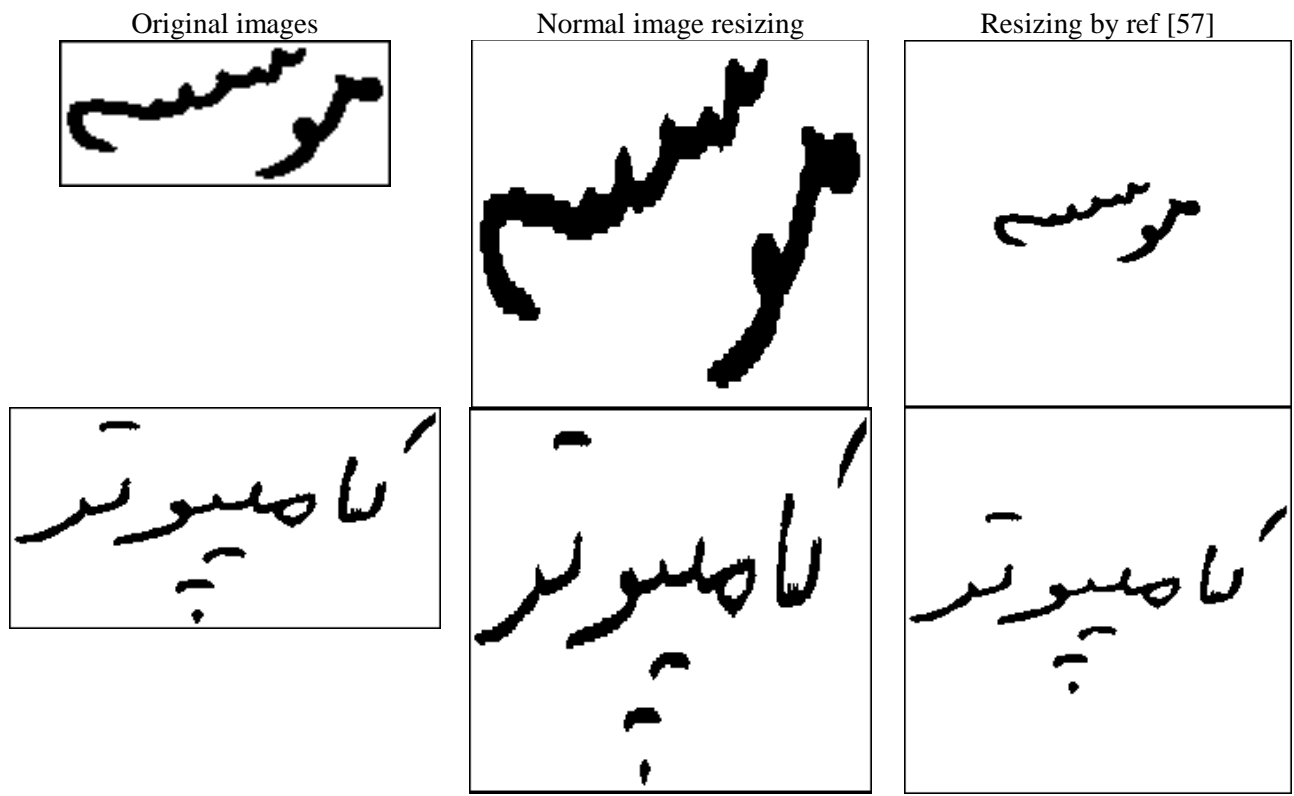

Fig 4. Column on the left: original image, Column in the middle: Resized image by normal scaling, and Column on the right: Resized image by [57].

## 4. Experimental results

The main goal of this present study is trying to answer fundamental question in document recognition. Using CNNs, we try to see whether color modalities of handwritten digits and words affect their recognition accuracy or speed. It should be emphasized that providing novel techniques for improving accuracy of handwritten word and digit recognition are not the objectives of these experiments. Two sets experiments are conducted: isolated digit recognition and handwritten word recognition. All our experiments run on a machine with Intel® core i3-6300 CPU @3.70GHz, 16GB RAM, and NVidia® 1060Ti 6GB GPU. We implement our experiments using PyTorch® framework installed on Microsoft® windows10.

### 4.1. Results for isolated handwritten digit recognition

In this part, we apply similar architecture (see Fig 1) for recognizing Persian handwritten digit database. In three image modalities, proposed CNN trained by Stochastic Gradient Descent (SGD) algorithm using empirical optimized parameters. The confusion matrixes for TC, GL, and BW images are shown in Tables 8, 9, and 10 respectively.

Table 8. The confusion matrix on TC images[52].

|   | 0 | 1 | 2 | 3 | 4 | 5 | 6 | 7 | 8 | 9 |
|---|---|---|---|---|---|---|---|---|---|---|
| 0 | **1983** | 0 | 5 | 1 | 0 | 2 | 2 | 3 | 1 | 3 |
| 1 | 0 | **1972** | 1 | 0 | 1 | 0 | 12 | 10 | 0 | 4 |
| 2 | 0 | 2 | **1864** | 78 | 39 | 1 | 10 | 5 | 1 | 0 |
| 3 | 0 | 0 | 60 | **1883** | 47 | 7 | 1 | 2 | 0 | 0 |
| 4 | 0 | 22 | 32 | 87 | **1822** | 8 | 5 | 15 | 4 | 5 |
| 5 | 4 | 8 | 1 | 7 | 14 | **1912** | 8 | 25 | 10 | 11 |
| 6 | 0 | 5 | 13 | 1 | 8 | 23 | **1412** | 507 | 14 | 17 |
| 7 | 0 | 14 | 56 | 13 | 120 | 109 | 288 | **1384** | 4 | 12 |
| 8 | 4 | 5 | 3 | 0 | 0 | 48 | 10 | 6 | **1904** | 20 |
| 9 | 1 | 10 | 1 | 0 | 5 | 57 | 16 | 25 | 7 | **1878** |

Table 9. The confusion matrix on GL images[52].

|   | 0 | 1 | 2 | 3 | 4 | 5 | 6 | 7 | 8 | 9 |
|---|---|---|---|---|---|---|---|---|---|---|
| 0 | **1992** | 3 | 0 | 0 | 2 | 3 | 0 | 0 | 0 | 0 |
| 1 | 0 | **1984** | 0 | 0 | 0 | 0 | 3 | 8 | 3 | 2 |
| 2 | 0 | 3 | **1928** | 52 | 8 | 0 | 0 | 4 | 4 | 1 |
| 3 | 0 | 0 | 55 | **1924** | 8 | 4 | 0 | 7 | 2 | 0 |
| 4 | 0 | 3 | 20 | 18 | **1932** | 3 | 1 | 10 | 4 | 9 |
| 5 | 0 | 5 | 0 | 0 | 9 | **1963** | 1 | 16 | 4 | 2 |
| 6 | 3 | 3 | 5 | 0 | 3 | 1 | **1521** | 455 | 0 | 9 |
| 7 | 0 | 26 | 34 | 15 | 13 | 48 | 100 | **1750** | 4 | 10 |
| 8 | 0 | 7 | 0 | 1 | 0 | 4 | 3 | 2 | **1976** | 7 |
| 9 | 0 | 9 | 0 | 0 | 4 | 42 | 7 | 13 | 1 | **1924** |

Table 10. The confusion matrix on BW images[52].

|   | 0 | 1 | 2 | 3 | 4 | 5 | 6 | 7 | 8 | 9 |
|---|---|---|---|---|---|---|---|---|---|---|
| 0 | **1989** | 2 | 2 | 0 | 3 | 4 | 0 | 0 | 0 | 0 |
| 1 | 0 | **1987** | 2 | 0 | 0 | 0 | 4 | 7 | 0 | 0 |
| 2 | 0 | 2 | **1937** | 36 | 17 | 0 | 1 | 5 | 2 | 0 |
| 3 | 0 | 0 | 56 | **1925** | 10 | 5 | 1 | 3 | 0 | 0 |
| 4 | 0 | 2 | 10 | 13 | **1950** | 3 | 0 | 13 | 3 | 6 |
| 5 | 0 | 5 | 0 | 1 | 7 | **1974** | 3 | 5 | 2 | 3 |
| 6 | 0 | 1 | 3 | 0 | 1 | 3 | **1562** | 424 | 0 | 6 |
| 7 | 0 | 10 | 7 | 8 | 6 | 29 | 56 | **1865** | 12 | 16 |
| 8 | 0 | 7 | 0 | 0 | 1 | 8 | 2 | 1 | **1972** | 9 |
| 9 | 0 | 13 | 0 | 0 | 4 | 40 | 3 | 8 | 0 | **1932** |

In addition, for better comparison among three image modalities, recognition accuracy for each isolated digit in TC, GL, and BW modalities are shown in Figure 5. As shown in this figure recognition accuracy for BW images are higher than two other image modalities, but these differences are not significant. Validation accuracy and loss values in three color modalities with respect to different epochs are shown in Figure 6. As shown these figures network, converges in BW images are faster than two other image modalities. Finally the training time for three color modalities are shown in Table 11. As shown in this table, in the training phase recognizing handwritten digits in BW modality is more efficient than the two other modalities.

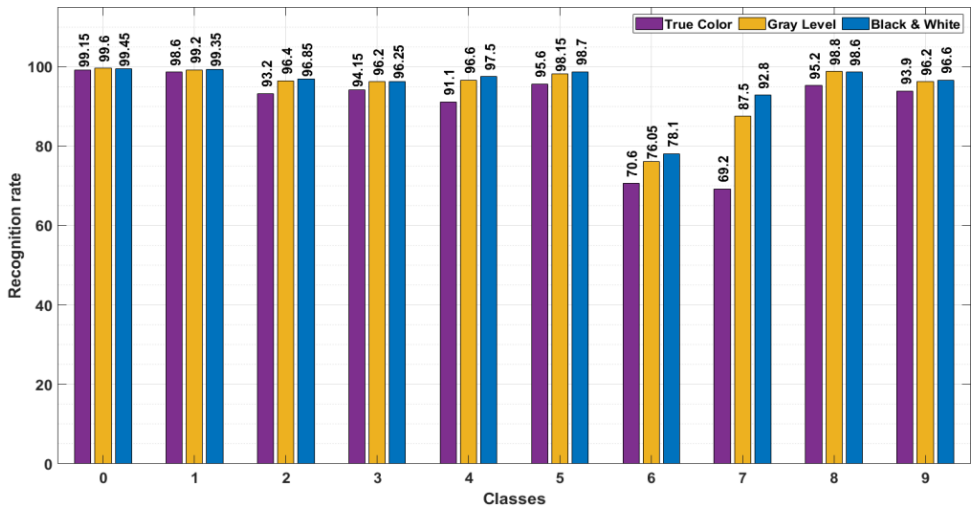

Fig 5. Overall recognition accuracy for each class (isolated digit) in three image modalities[52].

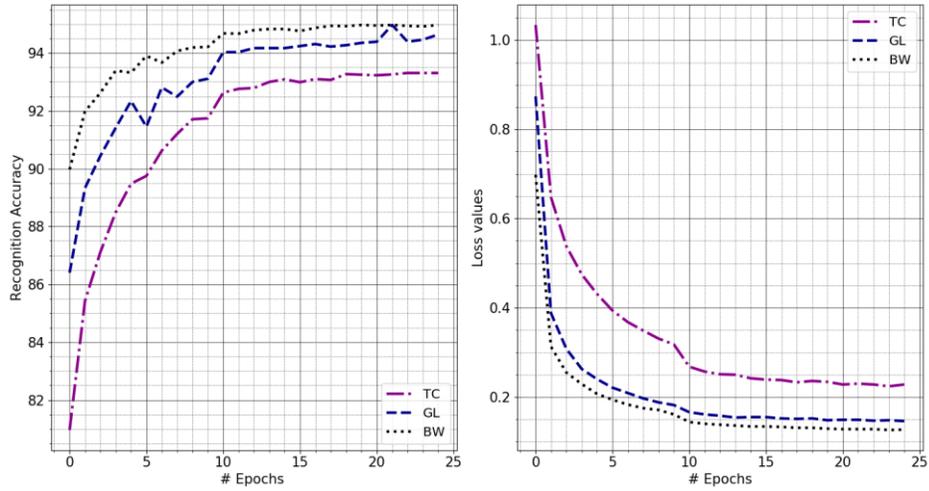

Fig 6. Performance of the proposed model with respect to different epochs during training in three different color modalities, (left) validation accuracy versus number of epochs, (right) loss values versus number of epochs for validation set[52].

Table 11. Runing times for training CNN (in seconds)[52].

| Color modalities | Training times (in seconds) |
| --- | --- |
| True Color(TC) | 68016.6 |
| Gray Level(GL) | 58705.5 |
| Black and White(BW) | 52261.0 |

### 4.2. Results for handwritten word recognition

As the second set of the experiments, samples of database [19] which consist of 62,500 words are selected, and then divided into 70% (0.7*62,500 = 43,750) for training, 15% (0.15*70,000 = 9,375) for validation and 15% (0.15*70,000 = 9,375) for testing. Figure 7 show validation accuracy and loss values in different epochs in three color modalities. As shown in this figure, similar to isolated digit (previous subsection) recognition accuracy and network convergence in BW images is outperformance than two other modalities but this difference isn't meaningful.

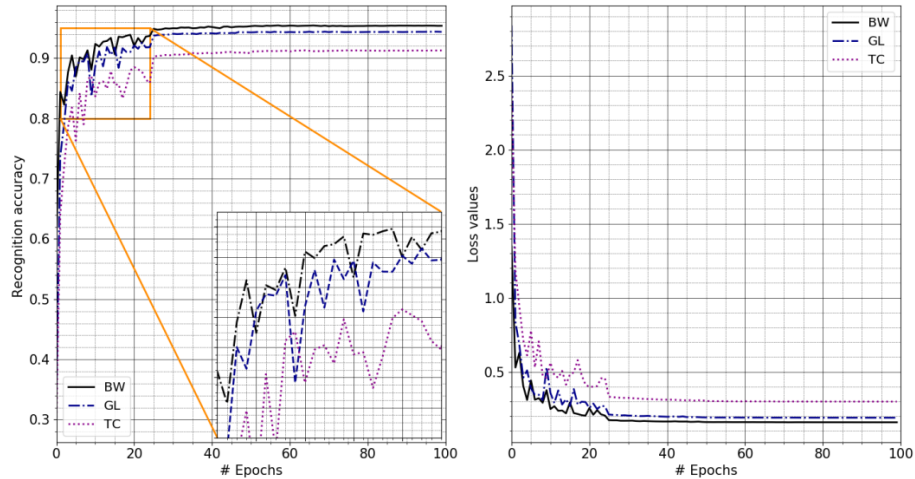

Fig. 7. Performance of the proposed model with respect to different epochs during training in three different color modalities, (left) validation accuracy versus number of epochs, (right) loss values versus number of epochs for validation set.

In Table 12, the performance of the fine-tuned network is compared to AlexNet with respect to training time and recognition accuracy in three color modalities. As shown in this table, compared to a backbone network (AlexNet), the training time was significantly reduced, while the network accuracy didn't decreased significantly in all three color modalities. Although the purpose of this paper is not to improve the accuracy of the existence Persian word recognition system, however compared to the most of previous works, the proposed method provides much higher results (see Table 13).

Table 12: Comparison of the performance of the best fine-tuned AlexNet and AlexNet (shown in Fig 3)

| Network | Total parameters | Training times (minutes) | | | Accuracy (%) | | |
|---|---|---|---|---|---|---|---|
| | | BW | GL | TC | BW | GL | TC |
| Alexnet | 57,515, 965 | 406 | 464 | 758 | 95.7 | 95.6 | 93.2 |
| Fine-tune Alexnet | 12,195,529 | 203 | 218 | 317 | 95.4 | 94.4 | 91.2 |

Table 13. Compare the performance of the proposed method with existing holistic methods in Persian/Arabic database

| Ref. | Method | Dataset | No. of classes | Accuracy (%) | | |
|---|---|---|---|---|---|---|
| | | | | TC | GL | BW |
| [20] | Heuristic | IRANSHAHR | 100 | ---- | ---- | 78.8 |
| [21] | Image gradient, classifier fusion | IRANSHAHR | 200 | ---- | ---- | 89.0 |
| [22] | Fuzzy vector quantization, HMM | IRANSHAHR | 198 | ---- | ---- | 67.7 |
| [23] | Structural based feature, discrete HMM, re-ranking | IFN/ENIT | 411 | ---- | ---- | 89.2 |
| [25] | Statistical and structural features, discrete HMM | IFN/ENIT | 411 | ---- | ---- | 87.9 |
| [26] | Right to Left, Left to Right HMM | IRANSHAHR | 200 | ---- | ---- | 84.4 |
| [14] | Image gradient, discrete HMM | IRANSHAHR | 198 | ---- | ---- | 68.8 |
| [27] | Chain code, HMM | IRANSHAHR | 198 | ---- | ---- | 69.0 |
| [28] | Statistical features, Support Vector Machine | IRANSHAHR | 198 | ---- | ---- | 80.7 |
| [29] | Connectionist Temporal Classification, bi-LSTM, CNN | Ref[19] | 125 | ---- | ---- | 98.1 |
| --- | **Modified AlexNet(Proposed network)** | **IRANSHAHR** | **503** | ---- | ---- | **89.4** |
| --- | **Modified AlexNet(Proposed network)** | **Ref** [19] | **125** | **91.2** | **94.4** | **95.4** |

5. Conclusion and future works

Unlike the retina of some animals like cats and dogs which possess solely blue and green cones, humans have color vision due to the existence of red, blue and green cones in the retina[1]. Color information plays an important role in human vision for recognition of elements such as shape, texture, and object recognition in daily life [2]. Similarly color as a rich feature or important descriptor is the most important part in many computer vision applications such as image super resolution[3], image segmentation[4], medical image analysis [5], remote sensing [6], image retrieval[7] and etc. In this paper we conducted a case study on Persian handwritten recognition in three different color vision modalities. Using CNNs, as eye simulators, we investigated to see whether color modalities of handwritten digits and words would affect their recognition accuracy or speed. To the best of our knowledge, so far this question has not been answered, quantitatively, due to the lack of handwritten digit and words databases that have all three-color modalities of the handwritten digits. To answer this question, samples of Persian isolated digits and words provided by a new comprehensive handwritten Persian database[19] in three different color modalities are selected and they are divided into training, validation, and testing sets. In general, experimental showed there are no significant difference for recognition accuracy in different color models in both isolated digits and words. Also, comparisons of training times in three color modalities show that recognition of handwritten digits and words in BW modality using CNN is much more efficient than two other color modalities. In the future study, we would like to investigate the effects of the color modalities on handwritten recognition of other script such as: Latin, Arabic, Chinese, and etc.